\documentclass{article}

    \PassOptionsToPackage{numbers, compress}{natbib}

    \usepackage[preprint]{neurips_2024}

\usepackage[utf8]{inputenc} 
\usepackage[T1]{fontenc}    
\usepackage{hyperref}       
\usepackage{url}            
\usepackage{booktabs}       
\usepackage{amsfonts}       
\usepackage{nicefrac}       
\usepackage{microtype}      
\usepackage{xcolor}         

\newcommand{\new}[1]{{#1}}

\title{Declare and Justify: Explicit assumptions in AI evaluations are necessary for effective regulation}

\author{%
  {Peter Barnett \hspace{2cm }Lisa Thiergart}\\
  Machine Intelligence Research Institute\\
  Berkeley, CA, USA 94704 \\
  \texttt{\{peter, \ lisa\}@intelligence.org} 
}

\begin{document}

\maketitle

\begin{abstract}
As AI systems advance, AI evaluations are becoming an important pillar of regulations for ensuring safety. We argue that such regulation should require developers to \textbf{explicitly identify and justify key underlying assumptions about evaluations} as part of their case for safety. We identify core assumptions in AI evaluations (both for evaluating existing models and forecasting future models), such as comprehensive threat modeling, proxy task validity, and adequate capability elicitation. Many of these assumptions cannot currently be well justified. If regulation is to be based on evaluations, it should require that AI development be halted if evaluations demonstrate unacceptable danger or if these assumptions are inadequately justified. Our presented approach aims to enhance transparency in AI development, offering a practical path towards more effective governance of advanced AI systems.
\end{abstract}

\section{Introduction}
The rapid pace of AI development has prompted demands for regulation to help safeguard against novel risks, including catastrophic risks~\cite{anderljung_frontier_2023}. This regulation should aim to prevent harms caused by malicious actors misusing AI systems, as well as large-scale accident risks caused by autonomous AI systems acting in misaligned ways~\cite{bengio_managing_2024, Bengio2024international, hendrycks2023overview}. 

Today’s frontier AI systems are not created by understanding and implementing specific capabilities; they are instead iteratively shaped through a training process that encourages instrumental capabilities to emerge. Consequently, AI developers do not know what their systems will be capable of until they test them — and sometimes not even then. As OpenAI CEO Sam Altman said about predicting capabilities~\cite{murgia2023openai}, “Until we go train that model, it’s like a fun guessing game for us.” 

Major AI developers~\cite{anthropic2023responsible, openAI2023preparednes, deepmind2024frontier} have put forward plans for safety based on AI evaluations  ~\cite{shevlane_model_2023, phuong_evaluating_2024} that attempt to assess a model’s capacity to facilitate dangerous activities such as hacking \cite{fang_llm_2024, fang_teams_2024}, bioweapons design \cite{soice_can_2023, pannu_prioritizing_2024}, and human manipulation \cite{salvi_conversational_2024, matz_potential_2024}. Governments are requiring that AI developers provide them access to models for testing~\cite{nistUSSafety, politicoRishiSunak}. Clearly, much depends on these evaluation efforts\new{, especially for avoiding potentially catastrophic risks}.   


But safety cases based on AI evaluations rest on many underlying assumptions about the scope and limitations of testing that may not have been adequately interrogated. Previous work discusses various limitations to AI evaluations~\cite{mukobi2024reasons, burden_evaluating_2024, hernandez-orallo_evaluation_2017, apolloresearchNeedScience}. In this paper, we \textbf{identify key assumptions} we argue should be stated and justified by developers as part of any safety plan or regulatory effort.

\section{Current AI evaluations workflow}
AI evaluations form a large component of AI developer safety plans, such as the Anthropic Responsible Scaling Policy~\cite{anthropic2023responsible}, the OpenAI Preparedness Framework~\cite{openAI2023preparednes}, and the Google Deepmind Frontier Safety Framework~\cite{deepmind2024frontier}. These seek to estimate whether current models have dangerous capabilities that could lead to catastrophic harm, and to predict if future models will. In this paper, we distinguish between existing and future models because the capabilities of future models can only be inferred from those of  existing models which can be directly interacted with. \new{These capability evaluations can include ``human uplift'' studies, where the capability being measured is the ability to assist humans at harmful tasks~\cite{ibrahim2024beyond, noauthor_AISafetyInstitute}. }

For evaluating \textbf{existing models}, the process is:
\begin{enumerate}
    \item Assess threat vectors via which the AI system could cause harm.
    \item Design proxy tasks which estimate the system’s ability to exploit these threat vectors.
    \item Attempt to get the model to do these proxy tasks.
    \item If a model can do these proxy tasks, trigger an action such as don’t release the model or don’t continue training the model without first resolving the risk. 
\end{enumerate}
An implicit assumption is made that if evaluators are unable to make a model perform well on the proxy tasks, then it is unlikely to have dangerous capabilities, and therefore will be safe to deploy.

Developers today acknowledge that some AI systems may not be safe to even create; for example, an inadequately secured model could be used to cause harm if stolen, or a model that has capabilities allowing it to break out of its containment could act autonomously. For forecasting and preventing risks from \textbf{future models}, it appears from the developer safety plans that the standard process is:
\begin{enumerate}
    \item Assess threat vectors via which \emph{future} AI systems could cause harm.
    \item Determine precursor capabilities that would appear before an AI system develops the actually dangerous capabilities.
    \item Design proxy tasks for these precursor capabilities.
    \item Attempt to get the model to do these proxy tasks.
    \item If a model ever displays these precursor abilities, stop development or deployment until sufficient precautions~\cite{nevo2024securing} are implemented.
\end{enumerate}

This approach makes the implicit assumption that there exists enough of a time and compute~\cite{sevilla2022compute} gap between reaching precursor capability and full required capability for evaluators to catch the precursors and stop further development.

\section{Key assumptions in AI evaluations}
For AI evaluations to provide justified confidence in a model’s lack of dangerous capabilities, several key assumptions must hold. Below, we explore a non-exhaustive listing of relevant assumptions.
\subsection{Evaluating existing models}
\paragraph{1. Comprehensive Threat Modeling: Have all the relevant threat vectors been considered?} Evaluators must adequately cover the space of dangerous capabilities the AI system could have that would allow it to cause harm. This requires threat modeling which covers \emph{all} exploitable threat vectors, including vectors evaluators didn’t consider but which the AI system might be capable of finding without detection.

Sufficient  justification for this assumption may be obtainable when the goal is to prevent harm via misuse by malicious actors. But this would require evaluators (potentially working with threat assessors and domain experts) to be correctly confident that they can find all threat vectors that malicious actors would be able to find. \new{AI developers have committed to working with domain experts as part of safety assessments~\cite{anthropic2023responsible, openAI2023openairedteaming}}. This could prove challenging, especially when considering well-resourced (potentially nation-state level) malicious actors.

\textbf{It may be substantially more difficult to justify this assumption when considering risks from autonomous AI systems}; for example, AI systems may notice threat vectors that humans do not.

\paragraph{2. Proxy Task Validity: Are the proxy tasks a good measure of the dangerous tasks?}  In order for the proxy tasks to be a good measure, it must be true that if an AI system does not succeed at a proxy task, the system cannot succeed at the more complex dangerous task. Therefore, success at the proxy must be a necessary requirement for success at the dangerous task. If there was an unanticipated alternative path to complete the dangerous task, the proxy would fail to provide a strong safety argument.

An assumption of proxy validity is potentially justifiable when considering misuse risk, although it requires the evaluators to understand the threat vectors as well as potential malicious actors do, which will likely require significant domain knowledge. It is not clear how to justify this assumption when considering risks from AI systems acting autonomously, because AI systems may be able to exploit threat vectors in inhuman ways that don’t require competence at the proxy tasks.

\paragraph{3. Adequate Capability Elicitation: Do evaluators adequately elicit model capabilities?} If evaluators are not able to elicit the AI system’s full capabilities~\cite{METRGuidelines}, this can result in false negatives, leading the evaluators to falsely believe a system to be incapable of a task. For example, if they did not use all possible post-training enhancements~\cite{davidson2023ai}. How can we assess whether the elicitation of a capability has been \emph{adequate}? This depends on the threat model being evaluated: 

\textbf{Misuse}: An assumption of adequate elicitation for the purposes of assessing potential misuse by malicious actors may be justifiable, although this may require asymmetrical provision of affordances: e.g. evaluators have access to non-safety-trained models and fine-tuning~\cite{casper_black-box_2024}, while public users have only monitored API access to the model. 

\textbf{Autonomy}: To assess if a model is capable of causing harm autonomously (i.e. misalignment risk), evaluators are not comparing against a human baseline, and must be able to elicit close to the “true limit” of a model’s capabilities. This effort could fail if there are methods for capabilities elicitation unknown to the evaluators, or if the model strategically underperforms on evaluations~\cite{van_der_weij_ai_2024}. It is currently unclear how an assumption of full elicitation of autonomous capabilities can be justified. 
\subsection{Forecasting future models}
\paragraph{1. Comprehensive Coverage of Future Threat Vectors: Have all the threat vectors relevant to the next iteration of models been considered?} This is similar to the assumption of comprehensive threat modeling when evaluating existing models, but with greater inherent uncertainty. Justifying  this assumption requires either a comprehensive mapping of possible future threat vectors or a robust argument that newly created models will not be capable of taking advantage of novel, unanticipated threat vectors. 

This assumption may be justifiable for misuse risks, although not at present given the nascent state of AI threat modeling. \new{This would require that future models do not have unexpected or unconsidered capabilities that malicious actors could learn about and use}. This assumption is not feasible to justify when considering risks from autonomous AI systems. Methods do not exist to predict if or when models will gain the ability to exploit threat vectors which evaluators may fail to consider.   

\paragraph{2. Validity of Precursor Capability Proxies: Are the proxy tasks a good measure of the precursor capabilities?} This assumption is again much the same as the assumption of proxy task validity for evaluating existing models, and faces similar difficulties. 

\paragraph{3. Necessity of Precursor Capabilities: Are the precursor capabilities necessary for the development of the dangerous capabilities?} If the precursor capabilities are not actually prerequisites for the dangerous capabilities, then AI developers may inadvertently create dangerous models because they did not observe the precursor capabilities.

Understanding of how capabilities arise in AI models is lacking, and there are not good methods for determining that certain capabilities will arise before others. While some capabilities may predictably arise before others (e.g. a model will likely be able to do basic programming before it is able to write complicated back-doored code), this assumption cannot, at present, be rigorously justified.

\paragraph{4. Adequate Elicitation of Precursor Capabilities: Do evaluators adequately elicit model precursor capabilities?} This assumption is again much the same as the assumption of adequate capability elicitation for evaluating existing models, and faces similar difficulties.

\paragraph{5. Sufficient Compute Gap between Precursor and Dangerous Capabilities: Is there a compute gap that is large enough to catch precursor capabilities before dangerous capabilities develop?} The gap must be large enough and the evaluation frequency high enough such that warning signs are caught before dangerous capabilities arise. Given the absence of methods to predict the size, or even the existence, of this gap, \textbf{this assumption cannot be robustly justified.} Precursor and dangerous capabilities may arise at the same time, such as if they result from an underlying third factor.  There are also no guarantees that there will not be sharp capabilities jumps~\cite{ganguli2022predictability,wei2022emergent}, or that models will not rapidly transition from having few to all of the precursor capabilities over a small increase in scaling. 

\paragraph{6. Comprehensive Tracking of Capability Inputs: Are all factors which lead to increased capabilities being tracked?} To forecast model capabilities, evaluators need to be tracking all the relevant factors that will change between existing models and more capable future models.This goes beyond just tracking the total compute used in training and should include architectural changes, data quality, different training setups, and other algorithmic changes. It may be feasible for evaluators to track the inputs into AI capabilities, especially if AI developers who understand these factors are required to honestly report on these.

\paragraph{7. Accurate Capability Forecasts: Are evaluators able to make accurate forecasts based on evaluations?} Probabilistic forecasts based on these evaluations must be accurate enough to reliably determine future model capabilities~\cite{changovernanceEvaluatingPredictions}. As well as tracking all relevant inputs contributing to capabilities, evaluators must also have a sufficiently accurate forecasting model for predicting how these inputs translate into future capabilities. 

Evaluators might gain some confidence in their predictions by establishing a good track record, but currently these track records do not exist. There may not be many generations of AI models before they become dangerous, making it challenging to establish a track record. Factors such as novel algorithmic progress could also disrupt these forecasts. 

\section{Implications for regulation}
AI regulation aimed at preventing catastrophic harm may amongst other components heavily  rely on AI evaluations. However, as discussed, gaining assurance of safety using AI evaluations relies on many underlying assumptions. \textbf{We propose that regulation based on AI evaluations should require AI developers to publish a list of the assumptions being made (e.g. the assumptions listed in this paper) and justify them, and these justifications should be subject to review by third party experts.} Justifying these assumptions is essential as part of a case for safety, and if the assumptions are not justified it is not appropriate to make inferences about a model’s capabilities beyond the specific tests. This proposal is intended as a practical measure to enhance transparency and assist regulators in determining whether AI development is safe. 

Evaluations can provide useful information about model capabilities, and should likely be performed even if the assumptions cannot be justified. But evaluations should not be used to argue that AI systems are safe in the absence of such justifications. AI evaluations should not provide a false sense of security, and rigorously listing assumptions may help alleviate this. 

We do not know exactly what AI regulation will look like, however regulation may be based on the capabilities of AI models; for example, models with certain capabilities may only be deployed with certain precautions, or AI developers may be required to argue that their systems are safe because they lack certain capabilities.

AI developers should explicitly state and justify the assumptions being made as part of an evaluations-based case for safety. These should be released for public scrutiny, as long as this itself would be safe (for example, it should not alert malicious actors to novel threat vectors). These assumptions and justifications should then be assessed by third-party experts. For example, if AI developers are required to publish safety and security protocols~\cite{weiner2024safe} which rely on evaluations, they should also include a comprehensive list of the assumptions being made, and how or whether these are justified.

Regulation could mandate that AI development (and certain deployments) should not continue if:
\begin{itemize}
    \item Evaluations reveal that a system has a sufficiently high probability of being dangerous, or that future systems will be dangerous before there are adequate security precautions. This could include triggering “red lines” or “yellow lines”–predefined specific capability thresholds beyond which AI development must stop or proceed only with extreme caution.  
    \item The AI developer does not provide a list of assumptions and justifications, or if these justifications are judged to be inadequate. When dealing with high-risk AI systems, development should not continue if the assumptions are not judged to hold with very high probability. 
\end{itemize}

\section{Conclusion}
In this paper, we have discussed the role of AI evaluations for avoiding catastrophic risks. We have identified key assumptions in safety cases based on AI evaluations, both for evaluating existing models and for forecasting future models. If AI regulation is to be based on such evaluations, it should require AI developers to comprehensively identify and justify these assumptions. This can offer regulators much needed transparency into determining whether AI development is safe. AI development should then be halted if evaluations reveal unacceptable dangers, or if the underlying assumptions cannot be adequately justified. 

\bibliographystyle{unsrt}

\bibliography{main.bib}

\end{document}